# Automatic driving lane change safety prediction model based on LSTM

Wenjian Sun[1]
Electronic and Information Engineering
Yantai University
Yantai, Shandong
swjhuman@gmail.com

Jingyu Xu[3]
Computer Information Technology
Northern Arizona University
Flagstaff, Arizona, USA
jyxu01@outlook.com

Linying Pan[2]
Information Studies
Trine university
Phoenix, Arizona, USA
panlinying2023@gmail.com

Weixiang Wan[4]
Electronics & Communication Engineering
University of Electronic Science and Technology of China
ChengDu, China
danielwanwx@gmail.com

Yong Wang[5]
Information Technology
University of Aberdeen,
Aberdeen, United Kingdom
fredia4jane@gmail.com

*Abstract*：Autonomous driving technology can improve traffic safety and reduce traffic accidents. In addition, it improves traffic flow, reduces congestion, saves energy and increases travel efficiency. In the relatively mature automatic driving technology, the automatic driving function is divided into several modules: perception, decision-making, planning and control, and a reasonable division of labor can improve the stability of the system. Therefore, autonomous vehicles need to have the ability to predict the trajectory of surrounding vehicles in order to make reasonable decision planning and safety measures to improve driving safety. By using deep learning method, a safety-sensitive deep learning model based on short term memory (LSTM) network is proposed. This model can alleviate the shortcomings of current automatic driving trajectory planning, and the output trajectory not only ensures high accuracy but also improves safety. The cell state simulation algorithm simulates the trackability of the trajectory generated by this model. The research results show that compared with the traditional model-based method, the trajectory prediction method based on LSTM network has obvious advantages in predicting the trajectory in the long time domain. The intention recognition module considering interactive information has higher prediction and accuracy, and the algorithm results show that the trajectory is very smooth based on the premise of safe prediction and efficient lane change. And autonomous vehicles can efficiently and safely complete lane changes.

*Keywords*：Prediction model; Safe lane change; LSTM; Deep learning

I. INTRODUCTION

In recent years, automatic driving has attracted wide attention around the world, and the public believes that automatic driving can improve traffic safety and reduce traffic accidents. In addition, it can improve traffic fluency, reduce congestion, save energy, and improve travel efficiency. In the relatively mature automatic driving technology, the automatic driving function is divided into several modules: perception, decision-making, planning and control, and a reasonable division of labor can improve the stability of the system. The higher the level of autonomous driving, the heavier the responsibility of the decision-making level, because the autonomous vehicle needs to have a very good sense of the surrounding environment, so that it can make judgments without driver intervention, so that the decision task is sent to the planning and control level[1-2]. Therefore, some self-driving manufacturers have conducted large-scale road tests, such as Google self-driving cars and Apple self-driving cars. However, due to the complexity of the traffic system, there are more or less some safety problems in the current automatic driving, leading to a series of accidents in the test of these automatic driving vehicles, the main reason is that the automatic driving algorithm inside the automatic driving vehicle is not enough to cope with the dynamic change of the traffic environment[3]. According to research, nearly one-third of all traffic accidents are caused by unsafe lane changes.

The human lane change execution model using machine learning is a data-driven model whose parameters need to be determined by training a large amount of lane change execution data. At present, there are few researches on human lane change trajectory planning using machine learning. Some relevant researchers apply k nearest neighbor algorithm to lane change trajectory planning, but the amount of data used in this model is very limited. Considering that existing machine learning algorithms can only predict the position of lane changing vehicles, DING Chenxi constructed a two-layer BP neural network to make real-time prediction of lane changing vehicles. This model learns NGSIM data and expands the data volume based on the existing research. However, lane change data is a kind of time series, and the above two machine learning methods are only a single copy of the position of the vehicle in a certain state, and do not take into account the connection between the lane change data of each planned step during lane change. On this basis, XIE Dongfan and other researchers built LSTM neural network to predict the lane change trajectory of vehicles, and achieved an accuracy of more than 99% for human lane change trajectory learning. Although these studies are all about the learning of human lane change behavior, they fail to take into account the safety problems generated in the process of lane change. However, rule-based lane change models do not provide uniform mathematical descriptions of lane change trajectories, resulting in a large number of trajectory curve equations, such as polar coordinate polynomial trajectories and quintic polynomial trajectories[4-6]. In addition, there is still no conclusion on which equation can best describe the lane change trajectory curve, and the current kinematic model describing the lane change process also has large errors. Considering the shortcomings of the existing methods, this paper modified the LSTM neural network and proposed a new lane change trajectory planning model for automatic driving based on a safety-sensitive improved long short-term memory network[7].

## II. LSTM NEURAL NETWORK LANE CHANGE MODEL

### A. LSTM structure

The LSTM model realizes the dynamic change of time scale under fixed parameters through self-cycling inside the cell. The data of vehicle driving behavior contains time-continuous information about vehicle driving behavior, so using LSTM model to model lane change intention recognition can achieve good simulation results. In this paper, long and short term memory neural network is used to train the safe lane change prediction model of autonomous vehicles, and long-term storage of safe lane change memory of autonomous driving is realized[8-10]. Besides, the long and short term memory (LSTM) network is trained and tested by using high-dimensional data set, and the information extracted by vehicle-to-vehicle (V2V) technology is combined. Using proven LSTM to predict the motion trajectories of surrounding vehicles. Based on the predicted trajectory information of the surrounding vehicles, a local path planning algorithm based on risk assessment and prevention is proposed.

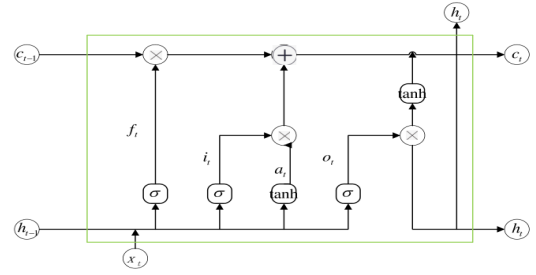

Figure. 1. LSTM model structure

The core structure of the LSTM network is composed of several cell structures in the above figure, and each cell structure has three control gates: forgetting gate, input gate, and output gate. The forget gate, which controls whether to forget, controls whether to forget the hidden cell state of the previous layer with a certain probability.

$$f_t = \sigma(w_f \cdot [h_{t-1}, x_t] + b_f) \quad (1)$$

Where δ represents the sigmoid function; $w_f$ represents the weight of the forgotten gate; ht-1 indicates the hidden state at t-1 moment. $x_t$ represents input data; bf represents the deviation of the forgetting gate. The input gate determines what information is updated to the cell state. it consists of two parts, the first part uses the sigmoid activation function, output as it; The second part uses the tanh activation function, which outputs at. The two results are then multiplied to update the cell status. The mathematical expression is:

$$i_t = \sigma(w_i \cdot [h_{t-1}, x_t] + b_i) \quad (3)$$

$$a_t = \tanh(w_c \cdot [h_{t-1}, x_t] + b_c) \quad (4)$$

Where δ represents the sigmoid function; tanh stands for tanh function; $w_t$ indicates the input gate weight; $w_c$ represents information about the candidate cell state that will be updated to the cell state; ht-1 indicates the hidden state at t-1 moment. $x_t$ represents input data; b represents the deviation of the forgetting gate; $b_c$ indicates candidate cell state deviation.

Then, the updating process of LSTM cell status is as follows:

$$c_t = f_t * c_{t-1} + i_t * a_t \quad (5)$$

The output gate outputs the information selectively and with the participation of the cell state:

$$o_t = \sigma(w_o \cdot [h_{t-1}, x_t] + b_0)$$

$$h_t = o_t * \tanh(c_t) \tag{6}$$

$o_t$ is the output of the output gate. $w_0$ is the output gate weight; $h_{t-1}$ indicates the hidden state at t-1 moment. $h_t$ indicates the hidden state at time t. $x_t$ represents input data; $b_0$ represents the candidate cell state deviation.

### B. Lane change planning model

Table1. Digital image technology application scenario case

Since the acceleration is required to change continuously during vehicle driving, the polynomial curve adopted by the lane change track of automatic driving should not be less than 3 times. In order to avoid too complicated parameter solving caused by too high number of polynomial curves, the polynomial curve is determined as a cubic polynomial curve, whose expression is as follows:

$$y_n(x_n) = a_0 + a_1 x_n + a_2 x_n^2 + a_3 x_n^3 \tag{1}$$

Where: $a_0$, $a_1$, $a_2$,; Are parameters that need to be determined later; The longitudinal position of the lane changing vehicle; y is the lateral position of the lane changing vehicle. Determine the following parameters:

$$y_n(x_n) = \tan\theta_i x_n + \frac{3y_n^f - 2x_n^f \tan\theta_i}{(x_n^f)^2} x_n^2 + \frac{x_n^f \tan\theta_i - 2y_n^f}{(x_n^f)^3} x_n^3 \tag{2}$$

In the formula, θ is the heading Angle of the starting point of the planned step size.

### C. Collision avoidance algorithm

Gipps model is a classic vehicle safety distance model in the field, which can better fit the driving state between two following vehicles, but it regards the vehicle as a particle without considering the vehicle body length[11]. Therefore, based on the classic Gipps model, the body length is added to improve it as the constraint condition of the cubic trajectory curve. In the process of lane change, vehicles are affected by the surrounding vehicles in real time. In order to ensure the safety of lane change process, it is necessary to detect the driving state of surrounding vehicles in real time and predict the driving state of surrounding vehicles.

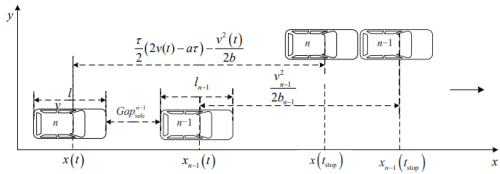

Figure. 2. Road safety distance model

Gipps model is to solve the current vehicle emergency stop, the rear car after the reaction time also take emergency stop action, so as not to crash the speed of the front car. In the classical Gipps model, the length of the car body is not taken into account, nor is the time-varying speed of the front and rear cars taken into account in the actual process of following[12].

### III. LSTM PREDICTION MODEL VERIFICATION RESULTS

The above experiments build two driver scenarios based on high-dimensional data sets for verification and evaluation. The proposed LSTM-MPC algorithm is compared with the MPC algorithm with constant velocity prediction and NIO network prediction. In the process of road planning, the driving safety and driving efficiency are compared according to RAI and collision avoidance speed[13].

Scenario 1: Active Lane Change When the car starts driving along lane 2, the target lane changes from lane 2 to lane 3 after a few miles. At the same time, there are three obstacles for vehicles near the self car in the 3rd lane, as shown in the picture below. The car needs to complete the lane change without a collision.

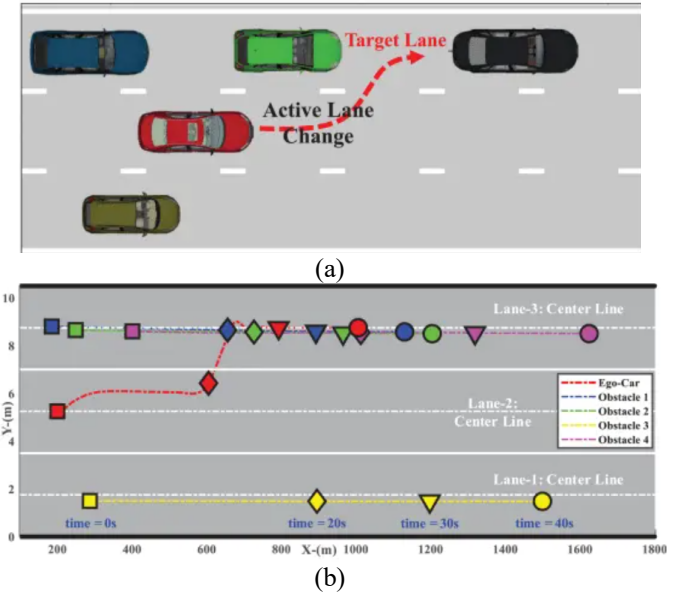

Figure. 3. Active lane change model

Horizontal position and RAI comparison between LSTM-MPC, NIO-MPC and conventional MPC implemented on the basic LSTM algorithm. The results in Figure 4(a) show that the path planning using LSTM-MPC has the lowest RAI than the other two methods, while the NIO-MPC method has a slightly higher RAI with a maximum error of 0.5 at 4 seconds. Figure 4(b) compares the lateral position of LSTM-MPC and conventional MPC during lane change. However, due to high forecast uncertainty, the lateral position of NIO-MPC is somewhat out of line with reality.

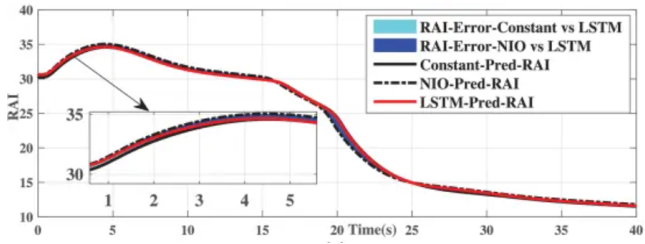

(a)

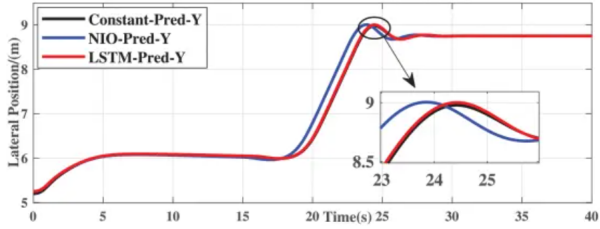

(b)

Figure. 4. Comparison of LSTM-MPC and MPC algorithm results

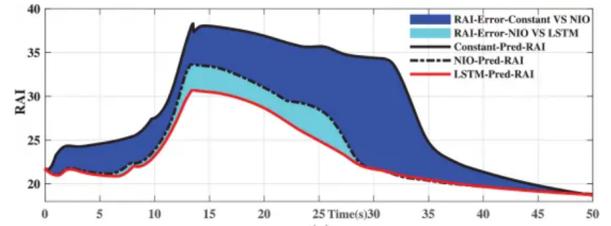

(a)

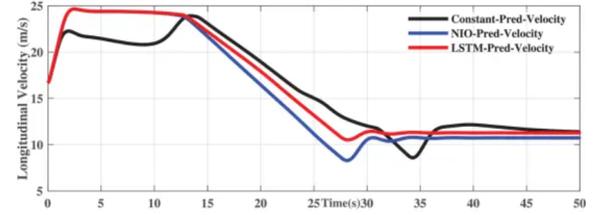

(b)

Figure. 5. Comparative analysis of LSTM and NIO

The second scenario is the emergency braking situation, and the track of the car is shown in the figure below. Blue and red represent the truck and the car, respectively, and the truck has a sudden deceleration time of about 15 seconds. The results showed that the self-driving car would also slow down to maintain a safe lane distance from the leading truck.

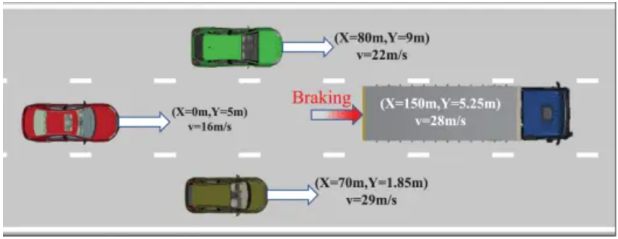

(a)

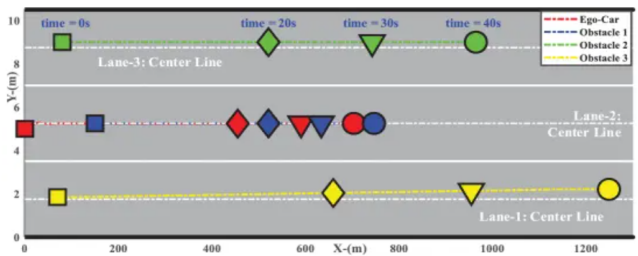

(b)

Figure. 4. Autopilot emergency braking model

The following figure shows the RAI of the three methods. The results show that in all kinds of collision avoidance methods, the self-driving vehicle will slow down as the main vehicle slows down. Using LSTM and NIO methods, the self-vehicle is decelerated in advance, and the RAI is improved compared with the constant method. However, due to the lower prediction accuracy of NIO, this leads to an unnecessarily larger deceleration than the other two methods.

Therefore, it can be seen that in various collision avoidance methods, the self vehicle will slow down with the deceleration of the main vehicle. Using LSTM and NIO methods, the self-vehicle is decelerated in advance, and the RAI is improved compared with the constant method. However, due to the lower prediction accuracy of NIO, this leads to an unnecessarily larger deceleration than the other two methods. In addition, the RAI value of LSTM-MPC algorithm is the lowest among the three methods, indicating that the self-vehicle has a higher safety level when using LSTM-MPC algorithm for path planning and path tracking[14]. By taking into account changes in the speed of surrounding vehicles during autonomous driving, the LSTM-MPC algorithm can be used to enable autonomous vehicles to travel with lower risk and shorter deceleration times.

## IV. CONCLUSIONS

In order to study the lane change technology of autonomous driving, an improved LSTM neural network for lane change trajectory planning is proposed in the field of autonomous driving. A deep learning lane change trajectory planning model is established from the perspectives of safety and efficiency, and the improved LSTM neural network model can improve the safety of the vehicle lane change process to a certain extent, so that the vehicle can use the rule-based algorithm to monitor and correct the safety of the trajectory while learning. In addition, it is verified that the comfort and efficiency of the improved LSTM neural network model are higher than that of the real trajectory, and the influence of lane-changing vehicles and surrounding vehicles on the lane-changing process is analyzed through Python simulation results[15-17]. Although the conventional lane change trajectory planning model has a high learning precision for the target trajectory during the execution of lane change, it ignores the changes of the surrounding environment of the vehicle, and the lane change vehicle cannot respond to unexpected situations, so there are still shortcomings in safety[18]. The experimental results of this paper fully demonstrate the path planning method based on risk assessment and risk prevention. The LSTM network is trained based on high-dimensional data sets and used to predict the motion trajectories of surrounding vehicles. By introducing the RAI index, risk assessment of the speed of surrounding vehicles is carried out to improve the driving safety of vehicles. The final comparative analysis shows that autonomous vehicles using LSTM-MPC always have the lowest RAI (highest safety) during autonomous driving under both active lane change and deceleration conditions. At the same time, the RAI comparison between this method and NIO method also shows that the higher the prediction accuracy of surrounding vehicles, the safer the self-vehicle will be in the process of path planning and tracking.

**Note: ①The authors of the references cited in the article should not be of one nationality only. They should be from three or more; ②It is not possible to have less than five references.**